\documentclass[10pt,a4paper]{article}

\usepackage[total={16cm,23cm}]{geometry}
\usepackage[]{url}
\usepackage{graphicx}

\title{Practical inventory routing:\\A problem definition and an optimization method}
\author{ M.J. Geiger$^1$ \hspace{4em} M. Sevaux$^{1,2}$\footnote{Corresponding author: \texttt{marc.sevaux@univ-ubs.fr}}\\[2ex]
\parbox{.45\textwidth}{
  \begin{center}
    $^1$Helmut Schmidt University\\
    Logistics Management Department\\
    Hamburg, Germany
  \end{center}}
\parbox{.45\textwidth}{
  \begin{center}
    $^2$Universit\'e de Bretagne-Sud\\
    Lab-STICC, Lorient, France
  \end{center}}
}

  \date{}

\begin{document}
\maketitle

\section{Introduction}
Many logistic activities are concerned with linking material flows among companies and processes. In such applications, we find a combination of quantity decisions, e.\,g.\ the amount of goods shipped (Inventory Management), and routing decisions as tackled in the area of Vehicle Routing. Clearly, both areas intersect to a considerable degree, complicating the solution of such problems. Recently, intensive research has been conducted in this context which is commonly refereed to as Inventory Routing Problems \cite{baita.ukovich.ea:98,bertazzi.savelsbergh.ea:08} (IRP). Several variants of the IRP can be found, ranging from deterministic demand cases to stochastic models.

From the practical point of view of the companies, reality is much more complex than a know demand and much more uncertain than a stochastic law. In fact, companies often have a partial knowledge of the demand over the planning horizon. Our observation of this phenomenon can be transformed in a new type of data, which we propose for further experimental investigations. We here assume that demand of the current period is known at the beginning of the period. Besides, we have an approximate overview of the demand over the 5 next periods, the 20 next periods and the 60 next periods. This overview is rather good (\emph{e.g.} it does not differ from reality by more that $\pm$10\%) but of course, we cannot predict with certainty what will happen the next periods.

The global objective of this work is to provide practical optimization methods to companies involved in inventory routing problems, taking into account this new type of data. Also, companies are sometimes not able to deal with changing plans every period and would like to adopt regular structures for serving customers.

As our work is a long term project, we are gradually going to develop our solution approach. In a first phase, we will focus on the Inventory Routing problem with a single product, deterministic known demand over a finite horizon. Contrary to \cite{archetti.bertazzi.ea:11}, we assume that the routing costs and the inventory costs are not comparable and therefore should be handled as two different objectives. To our knowledge, this is the first time that a bi-objective approach is considered for this problem.

\section{Problem definition}

Since our problem is somehow similar to \cite{archetti.bertazzi.ea:11}, we keep some of their notations in common. We consider a distribution network (usually a complete graph or a distance matrix) where a single product is shipped from a depot (denoted by 0) of unlimited capacity to a set $C=\{1,\cdots,n\}$ of customers over a time finite horizon $H$ of $p$ periods. A homogeneous fleet of trucks of capacity $K$ serves the customers (the number of trucks that can be used at every time period is not limited). Alternatively, a single truck can be used to do several tours over the same period. Each customer $i$ has a maximum capacity $U_i$ and an initial inventory level $I_{i0}$. The goal is to minimize two cost functions, namely the total routing costs (the sum of routing costs in each period) and the total inventory costs (the sum of the inventory levels at the end of each period for all customers).

\section{Proposed methodology}

Since this work is conducted in order to be used in practice, we have decided to develop different policies that can be understood and used on an every-day-basis by companies. Of course, the order-up-to-level-policy is provided as a choice as well as the day-to-day policy. In between, many policies exist. To make them simple, we intend to use a frequency-policy which will serve a customer for several periods in a row. Each customer will have its own frequency of delivery. Hence we will have to search for the best vector of frequencies that will propose non-dominated solutions for our two objectives. Initial routing is determinated by saving heuristics and latter improved using the Record-To-Record Travel algorithm \cite{li.golden.ea:07}.

\section{Preliminary results}

The final goal of this work is to be able to solve instances up to 250 customers over a long term horizon (240 periods). Instances have been generated based on geographical data from \cite{christofides.mingozzi.ea:79} and with three types of demand evolution (constant, increasing and sinusoid). Files are available on the web\footnote{\url{http://www2.hsu-hh.de/logistik/research/irp/GS-irp.zip}} and the file format is described in \cite{sevaux.geiger:11}.

\begin{figure}[!ht]
  \begin{center}
    \includegraphics[angle=-90,width=.7\textwidth]{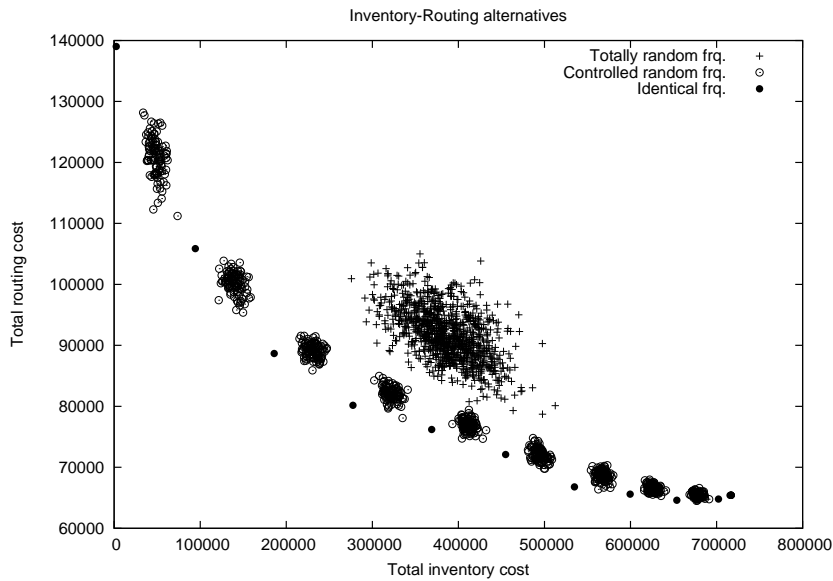}
  \end{center}
  \caption{Applying different policies}
  \label{fig:pareto1}
\end{figure}

Figure \ref{fig:pareto1} presents typical output of the frequency policies. The top left black dot is the day-to-day delivery policy, which clearly minimizes the inventory cost but has a routing cost which is important. Black dot on the bottom right represents the other alternative which apply the up-to-order policy. A large cloud of small crosses in the center results from a totally random frequency strategy. The black dots represent the solutions when all customers are served with the same frequency. To fill the gaps, a controlled random frequency (random frequencies but between two consecutive frequency values) is used and produces the results represented by white circles.

\begin{figure}[!ht]
  \begin{center}
    \includegraphics[angle=-90,width=.7\textwidth]{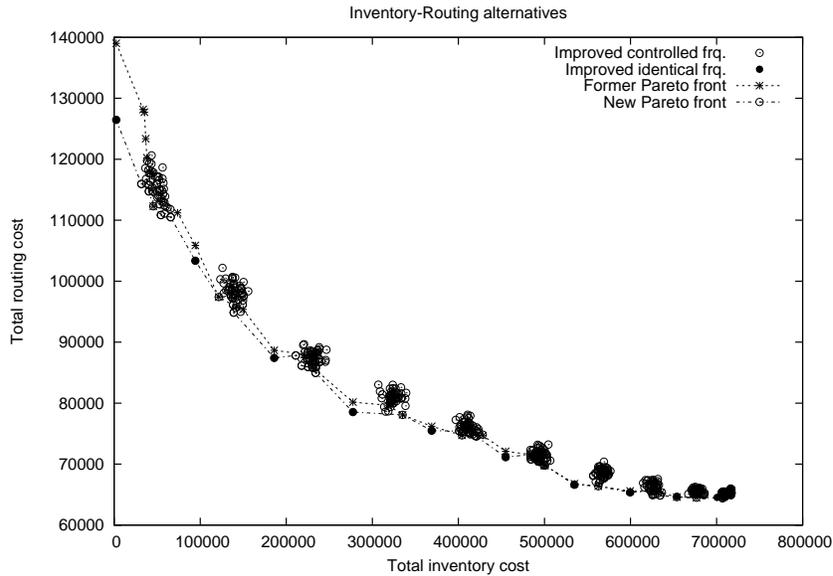}
  \end{center}
  \caption{Applying different policies, improved routing}
  \label{fig:pareto2}
\end{figure}

Using the work done in \cite{li.golden.ea:07}, routing cost can be improved. Figure \ref{fig:pareto2} shows the previous approximate Pareto front resulting from the left figure with an improved routing. With low frequencies, the routing cost can be reduced greatly.

\begin{figure}[htbp]
  \begin{center}
    \includegraphics[width=.6\textwidth]{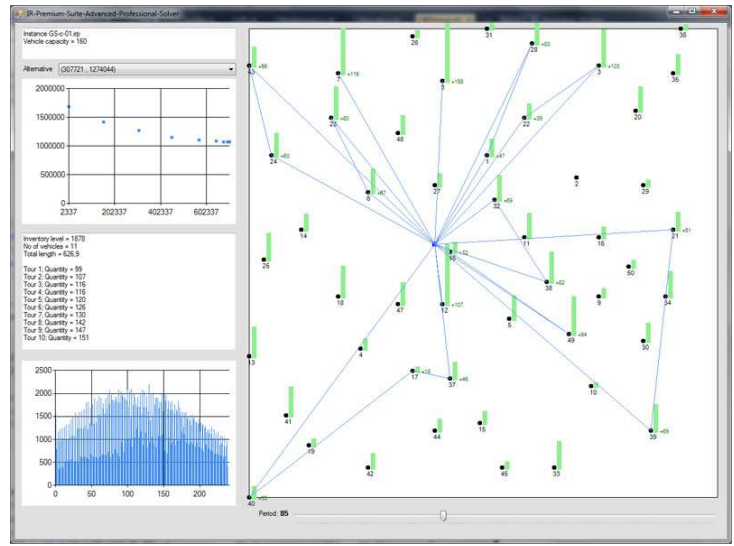}
  \end{center}
  \caption{Screenshot of the Inventory Routing Solver}
  \label{fig:screenshot}
\end{figure}

A test program has been developed and is used to test new and innovative strategies. Figure \ref{fig:screenshot} shows a typical screen shot of our solver. The upper part on the left gives the name of the instance and the vehicle capacity. Then below, the decision-maker is able to display the different alternatives computed by the software.

For the current alternative, and the current period, a text window gives the inventory level, the number of vehicle used and information on the tours. The box in the bottom left part represents the evolution of the total inventory over all periods. The large window on the right presents the current alternative and period routing. Green bars are the stock level at the customer location.

\section{Future work}

Several directions are now followed for the perspectives of our work. First, we are now improving our routing solver to avoid using an external software and create more intricated and embedded delivery strategies. Using the visual tool, we have already also noted some side effects that we could improve in our future research. A possible approach might be to change the synchronization of the customers at the beginning of deliveries or group customers depending on their geographical location.

\bibliographystyle{plain}

\end{document}